\documentclass{article}

\title{DDP-GCN: Multi-Graph Convolutional Network \\ for Spatiotemporal Traffic Forecasting}
\author{Kyungeun Lee$^{1}$, Wonjong Rhee$^{1,2 \ast}$ \\
    \small $^{1}$GSCST, Seoul National University, Seoul, Republic of Korea \\
    \small $^{2}$GSAI \& AIIS, Seoul National University, Seoul, Republic of Korea \\
    \small $^\ast$Corresponding author: Wonjong Rhee, wrhee@snu.ac.kr}
\date{}

\usepackage{graphicx}
\usepackage{amssymb}
\usepackage{lineno}
\usepackage{mathtools}
\usepackage{amsmath}
\usepackage{amsfonts}
\usepackage{subfig}
\usepackage{multirow}
\newcommand{\rulesep}{\unskip\ \vrule\ }

\usepackage{cite}
\usepackage{color,soul}

\providecommand{\keywords}[1]
{
  \small	
  \textbf{\textit{Keywords---}} #1
}

\usepackage{hyperref}
\begin{document}

\maketitle

\begin{abstract}
Traffic speed forecasting is one of the core problems in transportation systems. For a more accurate prediction, recent studies started using not only the temporal speed patterns but also the spatial information on the road network through the graph convolutional networks. Even though the road network is highly complex due to its non-Euclidean and directional characteristics, previous approaches mainly focused on modeling the spatial dependencies using the distance only. In this paper, we identify two essential spatial dependencies in traffic forecasting in addition to distance, \textit{direction} and \textit{positional relationship}, for designing basic graph elements as the fundamental building blocks. Using the building blocks, we suggest DDP-GCN (Distance, Direction, and Positional relationship Graph Convolutional Network) to incorporate the three spatial relationships into deep neural networks. We evaluate the proposed model with two large-scale real-world datasets, and find positive improvements for long-term forecasting in highly complex urban networks. The improvement can be larger for commute hours, but it can be also limited for short-term forecasting.
\end{abstract}


\section{Introduction}
\label{sec:intro}

Traffic forecasting is a crucial task for Intelligent Transportation Systems(ITS) \cite{tasci2018survey,lee2020short,veres2019survey,zhu2018itsbigdata,zhang2011itsdata}. Improving these forecasting systems is important for a wide range of applications, such as autonomous vehicles operations, route optimization, and transportation system management. In this work, we focus on the \textit{traffic speed forecasting}, which predicts the future traffic speeds for each segment of road using historical speed data. Accurate traffic speed forecasting can help prevent traffic congestion, shorten travel time, and reduce carbon emissions.

For a better prediction, recent deep learning studies have started to utilize not only the historical speed data but also the spatial information of the road networks. To manipulate spatial information into a format which can be used for deep networks like image-based CNNs, \mbox{\cite{ma2017image,polson2017deep,niu2019novel,zang2018long,wang2016traffic,cui2018deep,liao2018dest,shen2018research,liu2018short,lv2014traffic,pamula2018impact,wu2018hybrid}} simply unfold the road network and \mbox{\cite{yu2017sensors,ma2018capsule,duan2018improved,yao2019revisiting,chen2018exploiting,yao2018modeling}} use a grid-based representation resulting in map-like images. However, these image-like representations cannot fully capture the complex spatial relationships of traffic networks, such as driving directions and on-path proximity. To better understand the non-grid spatial characteristics of the traffic networks, recent works have started to employ the \textit{graph convolutional networks}(GCNs).

While most of the previous works are limited by using Euclidean distance as the only graph element for GCNs \cite{yu2017stgcn,chen2019gated,yu2019stunet,wang2019,pan2019urban}, some studies expand GCNs to include non-Euclidean dependencies.  \cite{lin2018predicting} provides one of the earliest GCN applications in traffic forecasting where graphs are defined by several types of non-Euclidean dependencies, such as total demand between stations, average trip duration, and demand correlation coefficient. \cite{li2017dcrnn,cui2018reachability} modified distance graph with non-Euclidean relationships, such as inflow/outflow and reachability. For bike demand forecasting, \cite{chai2018bike,geng2019stmgcn,geng2019multi} implemented multi-graph convolution based on the three types of graph elements, such as transportation connectivity and functional similarity, in addition to the distance. To adapt multi-graph convolution to traffic forecasting, we first need to understand which non-Euclidean dependencies are important for traffic networks.

\begin{figure}[ht!]
\centering
    \includegraphics[width=0.6\textwidth]{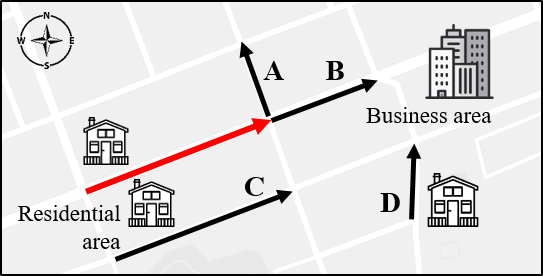}
    \caption{An example of the importance of non-Euclidean and directional characteristics of traffic networks.
    Target link is colored in red.
    }
    \label{fig:motivation}
\end{figure}

Figure \ref{fig:motivation} shows a simple example.
Here we want to figure out how the speed pattern of the target link(colored in red) is related to the other four links, $A$, $B$, $C$, and $D$. 
If we consider the \textit{distance} only, the directly neighbored links $A$ and $B$ might be the most related links to the target. However, due to the different \textit{driving direction}, $A$ might show quite a different speed pattern.
Compared to $A$, $C$ could have a more consistent speed pattern with the target, because $C$ shares the target's direction.
Additionally, $D$ could also share a similar speed pattern with the target because they are heading to the \textit{same area}. These properties might be more pronounced during commute hours. 

In order to utilize these concepts, we define two types of spatial dependencies in our work in addition to \textit{distance}: 
\textit{direction} and \textit{positional relationship}. 
Then, we propose a new type of traffic prediction network called DDP-GCN (Distance, Direction, and Positional relationship Graph Convolutional Network).
In DDP-GCN, the non-Euclidean characteristics of direction and positional relationship of the complex road networks are described through \textit{multi-graphs}. In previous studies \cite{chai2018bike,geng2019stmgcn,geng2019multi}, multi-graph convolution has been defined using undirected graphs consisting of links without direction information for the bike demand prediction task. However, the same approach cannot be directly applied to traffic forecasting on the directed graphs, especially consisting of links with \textit{direction} information.
In this work, we define multi-graphs based on link vectors and link directions. We also suggest partition filters for further improving the graph elements. A partition filter can be used to sub-divide each spatial graph element into multiple components with similar characteristics.
When evaluated on two large-scale real-world datasets that are highly complex urban networks, DDP-GCN outperformed the state-of-the-art baselines.

Our main contributions are in two-folds.
\begin{itemize}
    \item We identify non-Euclidean spatial relationships, \textit{direction} and \textit{positional relationship}, and propose to encode them using multiple graphs. We also suggest partition filters. 
    
    \item We propose a traffic forecasting network(DDP-GCN) which exploits the desired spatial dependencies effectively.
    This model is especially beneficial for long-term forecasting in a highly complex urban network, known as the most challenging problem.
\end{itemize}

The rest of this paper is organized as following.
Section \ref{sec:related_work} investigates how the spatial dependencies are represented in traffic forecasting. We also describe the graph convolutional networks and how they are used for traffic forecasting. Section \ref{sec:definitions} defines the key concepts and formulates the research problem. Section \ref{sec:methodology} elaborates our model DDP-GCN. Section \ref{experiment} reports the prediction performance and ablation test results for two real-world large-scale datasets. Section \ref{conclusion} provides the conclusions.
\section{Graph Convolutional Networks for Traffic Forecasting}
\label{sec:related_work}

The goal of traffic forecasting is to model spatiotemporal relationships and complex interplay of road networks. 
Traditional methods such as Auto Regressive Integrated Moving Average (ARIMA) and Kalman filters have mainly focused on temporal relationships under the time stationary process assumption. Later, their extensions were studied to include spatial relationships or to relax the stationarity assumption. For instance, \cite{ding2011forecasting,cheng2012spatio,pfeifer1980three,kamarianakis2005starima} enhanced the basic ARIMA model to spatiotemporal ARIMA (ST-ARIMA).

While the traditional methods and their extensions have been used with a great success, they are limited in that 
the methods do not have enough flexibility to model the traffic network with highly nonlinear and complex spatiotemporal characteristics~\cite{yu2017stgcn}. 
Instead, recent works have begun to apply deep learning approaches to traffic forecasting. Unlike the traditional models, Deep Neural Network (DNN) models have sufficiently large capacity and thus a potential for a large improvement.
In this section, we briefly summarize the fundamentals of applying DNN to the traffic data modeling, including input data representation and graph convolutional network.

\subsection{Representing Traffic Data for DNN}

\begin{figure}[t!]
    \centering
    \subfloat[]{
    \includegraphics[width=0.6\textwidth]{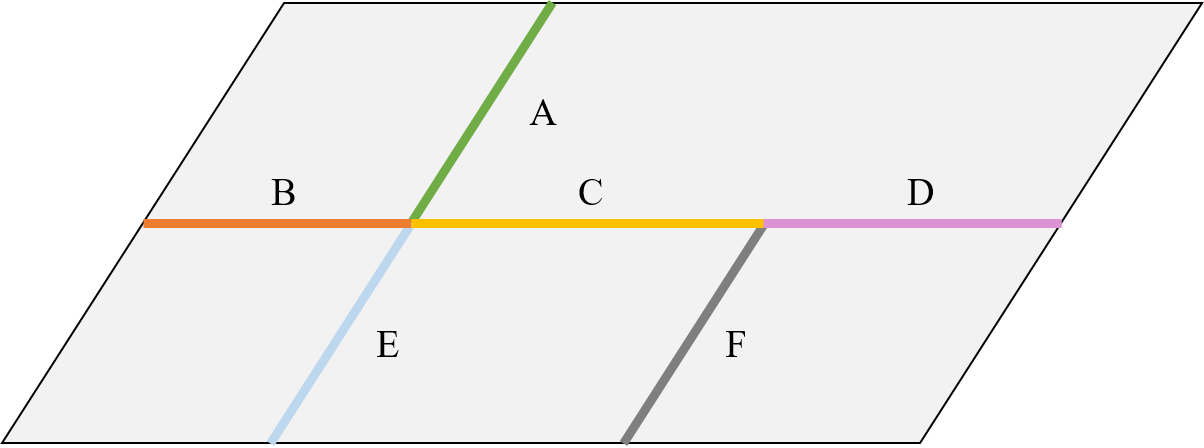}
    } \\ 
    \subfloat[]{
    \includegraphics[height=8em]{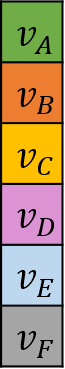}
    } \rulesep
    \subfloat[]{
    \includegraphics[height=8em]{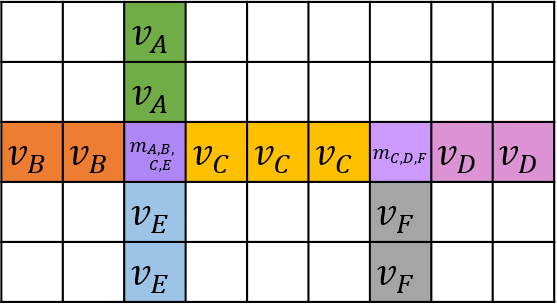}
    } \rulesep
    \subfloat[]{
    \includegraphics[height=8em]{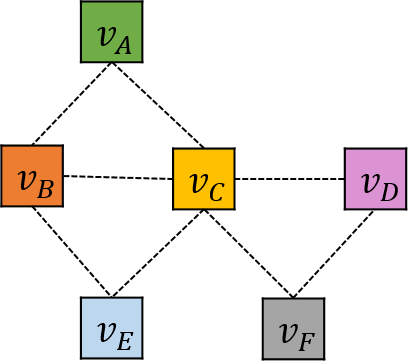}
    }
    \caption{Three ways of representing a road network as an input to deep neural networks. (a) Illustration of an exemplary road network with six road links.
    (b) Representation as a simple vector. $v_i$ is the value of speed or other value of interest for the link $i$. 
    (c) Image-like representation using a two-dimensional matrix. $m_{i_1, ..., i_N}$ is the average value of $\left\{ v_{i_1}, ..., v_{i_N}\right\}$.
    (d) Representation as a graph. The spatial relationship information contained in the graph is converted to a 6$\times$6 matrix where $(i,j)$ value represents distance, connectivity, or any other relational information between road links $i$ and $j$. 
    }
    \label{fig:spatial-rep}
\end{figure}

As the first step, we explain the most common techniques for representing road network data as the input to the DNNs. In theory, deep learning has the capability to learn any sophisticated spatiotemporal relationship within the data. In practice, however, the actual performance after training is heavily dependent on how the data is presented to the deep neural networks and therefore the data representation is an important matter. A brief summary of the three most common techniques are shown in Figure \ref{fig:spatial-rep}.

The first representation method is shown in Figure \ref{fig:spatial-rep}(b) where the road network data is simply organized into a vector form. Most of the initial DNN works followed this practice of \textit{stacked vector} \cite{ma2017image,polson2017deep,niu2019novel,zang2018long,wang2016traffic,cui2018deep,liao2018dest,shen2018research,liu2018short,lv2014traffic,pamula2018impact,wu2018hybrid}. 
While no clear rule has been established on how to stack the data into a single vector, human intuition has been utilized when possible. For instance, a circular road network was considered in \cite{ma2017image} and the stacking was performed over 352 links in a clockwise order. The vector stacking is a simple and flexible method, but it does not provide principled procedures of determining the representation. 

The second representation method is shown in Figure \ref{fig:spatial-rep}(c) where the road network data is organized into a two dimensional grid data. The \textit{2D-grid} representation, also known as grid-based map segmentation\cite{zhang2017deep,yu2017sensors} or grid map \cite{yao2019revisiting,chen2018exploiting,yao2018modeling}, has become popular because of its image-like representation. For image data, Convolutional Neural Networks (CNN) has been extremely successful. By representing road network data into a 2D-grid, it becomes possible to directly  apply the well developed CNN solutions to the traffic forecasting. \cite{yu2017sensors,ma2018capsule,duan2018improved,yao2019revisiting,chen2018exploiting,yao2018modeling} are examples where 2D-grid representation was adopted. The downside of 2D-grid lies in its inefficiency in representation. As shown in Figure \ref{fig:spatial-rep}(c), a vast majority of grid points do not correspond to any road link and zero needs to be inserted to indicate the absence of road link. The inefficiency can be aggravated when there is a need for a higher resolution. When each grid point corresponds to a large square area, for instance 30m$\times$30m, some of the grid points can include multiple road links where only one or average of all can be expressed. To mitigate the problem, the resolution needs to be sufficiently increased resulting in even worse efficiency in terms of the 2D matrix size for the representation. This can be a serious problem especially when a pair of road links heading to the opposite directions need to be represented. 

The last representation method is shown in Figure \ref{fig:spatial-rep}(d) where the road network data is represented as a \textit{graph}. 
A graph consists of nodes (vertices) and edges, and each node in Figure \ref{fig:spatial-rep}(d) corresponds to a road link in Figure \ref{fig:spatial-rep}(a). With graph representation, information can be represented in two different forms.
When simple per-node information needs to be represented, a stacked vector can be used as in Figure \ref{fig:spatial-rep}(b).
When complex pair-wise information needs to be represented, a $N\times N$ matrix can be used where $N$ is the number of nodes. 
The matrix is typically called \textit{adjacency matrix}, and 
its ($i$,$j$) element represents the pair-wise relationship between links $i$ and $j$.
Examples of complex pair-wise information include distance, connectivity, and other spatial relationship.
Graph representation has become popular in recent works \cite{yu2017stgcn,chen2019gated,yu2019stunet,wang2019,pan2019urban,li2017dcrnn,cui2018reachability}, because the $N \times N$ matrix is an efficient way of representing spatial relationship and because the resulting DNN models have shown a promising performance.  

\subsection{2D Convolution vs. Graph Convolution}

\begin{figure}[h]
    \centering
    \subfloat[2D Convolution]{
        \includegraphics[height=12em]{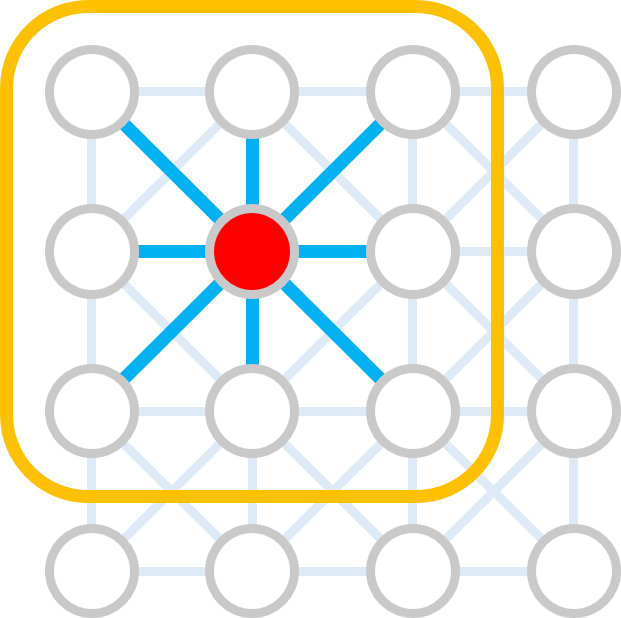}
    } \hspace{0.25cm}
    \subfloat[Graph Convolution]{
        \includegraphics[height=12em]{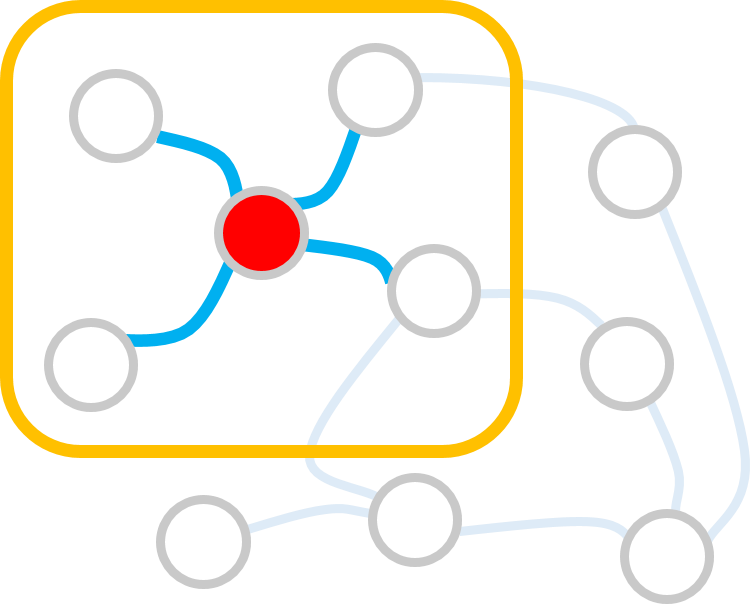}
    }
    \caption{
    Illustration of a typical 2D convolution and a graph convolution. For both, the information of the neighboring nodes defined as the nodes inside the filter (orange box) is used for processing the information of the node of interest (shown in red).
    (a) 2D convolution -  Defining filter shape and size is straightforward because the underlying data is well structured. All the neighboring nodes connected with blue lines are located inside the filter.  
    (b) Graph convolution - Unlike grid data, traffic network data is not well structured and a node's neighborhood size can vary. Instead of defining a small and common filter for all nodes, $N\times N$ matrix is used as the spatial filter where the varying neighborhood information is collectively represented in the matrix. In the column $j$ of the matrix, all the neighboring nodes connected with blue lines from node $j$ are represented with non-zero weight values while the others are marked with zero.
    (The figures were adapted from Figure 1 of \cite{wu2019survey}.)
    }
    \label{fig:gcn}
\end{figure}

Similar to the typical convolution operation on image data, graph convolution can be described as a weighted averaging over the neighborhood nodes \cite{wu2019survey} where graph locality is used as the inductive bias.
For image data as illustrated in Figure \ref{fig:gcn}(a), defining a neighborhood with a $3\times3$ filter is straightforward where the adjacent eight pixels are chosen as the neighbors. 
To adapt the image convolution operation to the graph convolution, we first need to define a proper neighborhood for applying a filter as the example shown in  Figure \ref{fig:gcn}(b). Unlike images, however, graphs have irregular and unordered data structures and a neighborhood cannot be simply defined as in the image convolution. Instead, the inter-node relationships need to be identified first. Typically domain knowledge is required, and usually proximity and on-path distance are used for defining a neighborhood within a road network.
As an example, we can consider the road network shown in Figure \ref{fig:spatial-rep} again. If we are interested in defining one-hop neighborhood, the graph convolution filter for learning the representation of link A will contain the links B, C, and E. 

\subsection{Brief History of Graph Convolutional Networks}
Graph Convolutional Network (GCN) was first introduced in \cite{bruna2013spectral} where spectral graph theory and deep neural networks were bridged.
Then \cite{defferrard2016} proposed ChebNet, which improved GCNs with fast localized convolution filters using Chebyshev polynomials. 
ChebNet implicitly avoided the computation of graph Fourier basis and
significantly reduced the computational complexity.
\cite{kipf2016semi} introduced 1stChebNet, as an extension of ChebNet.
1stChebNet not only provided a competitive performance for a variety of tasks
but also greatly reduced the computational cost by avoiding the eigenvalue decomposition that was required for calculating the Fourier basis.
Additional details on the history of GCN can be found in \cite{kipf2016semi,wu2019survey} with further explanations.

\subsection{Traffic Forecasting with GCN}
As GCNs heavily depend on the Laplacian matrix of a graph, it is crucial to determine a proper edge weighting that reflects the network geometry sufficiently well.
For traffic forecasting problems,
\cite{iyer2018urban,zhao2019t,zheng2019gman} define the road network as an undirected binary graph where each edge indicates if the two nodes are directly adjacent or not. \cite{yu2017stgcn,wang2019,yu2019stunet,pan2019urban} expand the road network graph as a weighted graph where each edge weight is determined to be inversely proportional to the physical distance or the travel time between the corresponding two nodes in the network.

Some of the recent studies have started to reflect other factors additional to the distance information. 
\cite{li2017dcrnn,cui2018reachability} modified the distance graph with additional relationships such as inflow/outflow and reachability.
For the case of bike demand forecasting, 
\cite{chai2018bike,geng2019stmgcn,geng2019multi} considered three types of graph elements including transportation connectivity and functional similarity, and implemented a multi-graph convolution as the sum of the individual operations.
For traffic flow prediction, multi-graph convolution was considered in \cite{lv2020temporal} recently. In the paper, four types of graphs were applied to address semantic pair-wise correlations among possibly distant roads. In this work, we newly define two non-Euclidean graph elements and suggest a partition filter that can be used to modify each graph element. Then, we investigate possible designs of multi-graph convolution for incorporating the newly developed graph elements. 
\section{Definitions and Problem Formulation}
\label{sec:definitions}

\begin{figure}[h!]
\centering
    \includegraphics[width=0.7\textwidth]{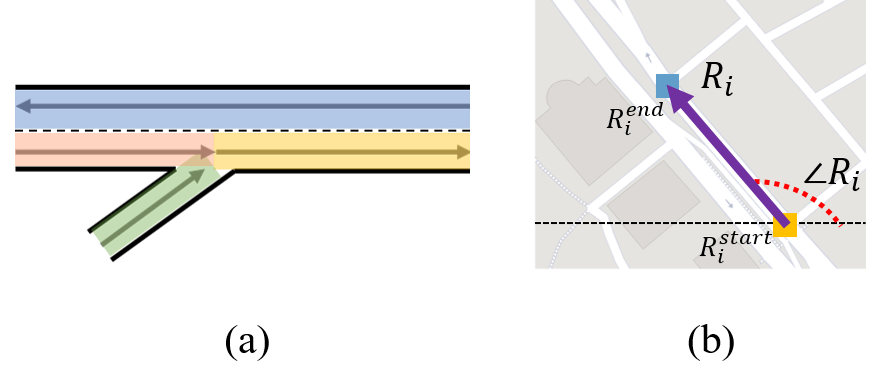}
    \vspace{-0.2cm}
    \caption{Defining links (a) Examples of four links shown in four different colors. (b) A link vector $R_i$, that represents a vector starting at $R_i^{start}$ and ending at $R_i^{end}$ in 2-D space, and its link direction $\angle{R_i}$.}
    \vspace{-0.2cm}
    \label{fig:link}
\end{figure}

In this section, we define the key concepts for modeling road traffics and formulate the problem.
Using the link concept from \cite{seo2017traffic}, we newly define link vector and link direction as below.
In general, a \textit{link} represents a road segment without an internal merge/diverge section, as shown in Figure \ref{fig:link}(a).

\vspace{0.3cm}
\textbf{Definition 1}: 
\textit{
The \textbf{link vector} $R_i$ of a link $i$ is defined as the difference between the end point 
$R_i^{end}$ and the start point $R_i^{start}$, and can be formulated as  
\begin{equation}
    R_i = R_i^{end} - R_i^{start}
    \label{def:linkvector}
\end{equation}
where $R_i, R_i^{start}, R_i^{end}\in\mathbb{R}^2$. For link vector $R_i$, its \textbf{link direction} $\angle{R_i}$ is defined as 
\begin{equation}
    \angle{R_i}=\arccos
    \left ( 
    \frac{R_i \cdot e_x}{\left \| {R_i} \right \|_2}
    \right ) 
    + 
    \left ( \frac{\text{sign}\left ( R_i \cdot e_y \right )+1}{2} \right ) \pi,
    \label{def:direction}
\end{equation}
where $e_x$ and $e_y$ are the unit vectors in the direction of the x-axis and y-axis, respectively.
}

Note that $\angle{R_i}$'s value could be between 0 and $2\pi$. An illustration of a link vector and link direction are shown in Figure \ref{fig:link}(b).

\vspace{0.3cm}
\textbf{Definition 2}:
\textit{
A \textbf{traffic network graph} is a weighted directed graph $\mathcal{G}=\{V,E,W\}$ representing a road network, where 
$V$ is the set of road links with $|V|=N$,
$E$ is the set of edges representing the connectedness among the road links, and 
$W \in \mathbb{R}^{N\times N}$ is a weighted adjacency matrix representing spatial inter-dependencies. 
}

Usually, the weighted adjacency matrix has $W(i,j)=0$ when the road links $i$ and $j$ are not connected. However, we will define new adjacency matrices later, where this property does not necessarily hold. It is also noted that $E$ is not used in our work because the connectedness is fully described by $W$. Finally, we formulate the problem as below. 

\vspace{0.3cm}
\textbf{Problem}: 
\textit{
    If a graph signal $X \in \mathbb{R}^{N\times 1}$ represents the traffic speed observed on 
    $\mathcal{G}$, 
    and $X^{(t)}$ represents the graph signal observed at $t$-th time interval,
    the traffic forecasting problem aims to learn a function $g(.)$ that 
    maps the $T'$ historical graph signals to 
    the $T$ future graph signals. For a given graph $\mathcal{G}$,
    \begin{equation}
        \left [ X^{(t-T'+1)}, ..., X^{(t)}\right ] \overset{g(.)} {\longrightarrow} \left [ X^{(t+1)}, ..., X^{(t+T)} \right ].
    \label{def:problem}
    \end{equation}
}

In general, $X$ can be of size $\mathbb{R}^{N\times P}$, where $P$ is the number of observed features for each link. 
Even though our dataset includes only the speed feature, i.e. $P=1$, all of our results are directly applicable to the problems with $P>1$.
\section{Proposed Model}
\label{sec:methodology}

Before explaining our methods, we briefly summarize the general graph convolution based on the approximation of 1stChebNet\cite{kipf2016semi} for a single directed graph $\mathcal{G}$. 
For a directed graph $\mathcal{G}=\{V,E,W\}$, 
1stChebNet generalizes the definition of a graph convolution as
\begin{equation}
    \theta \ast x \approx \theta \left ( I_N + D^{-1}W \right ) x,
    \label{def:gc}
\end{equation}
where the signal $x \in \mathbb{R}^N$, a scalar for every node,
the learnable parameters $\theta \in \mathbb{R}^N$, 
and the diagonal degree matrix $D \in \mathbb{R}^{N \times N}$ with $D_{ii}=\sum_j {W}_{ij}$.\footnote{
For a concise explanation, 
we will refer to graph convolution as defined above, 
however, it also can also be generalized to multi-dimensional tensors \cite{kipf2016semi,yu2017stgcn}.}

\subsection{Framework Overview}

\begin{figure*}[ht!]
    \centering
    \subfloat[Full Model]{
        \includegraphics[height=15em]{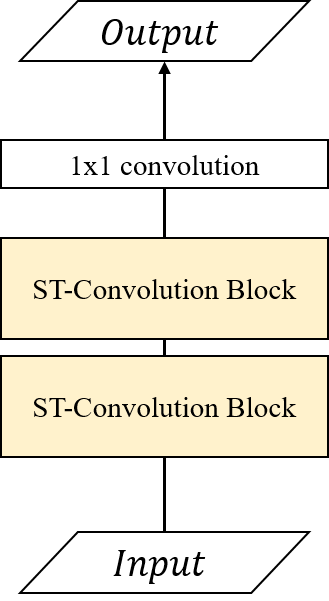}
    } \rulesep
    \subfloat[ST-Convolution Block]{
        \includegraphics[height=15em]{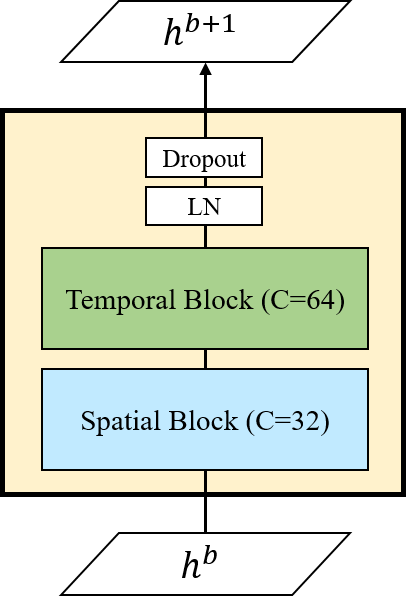}
    } \rulesep
    \subfloat[Temporal Block]{
        \includegraphics[height=15em]{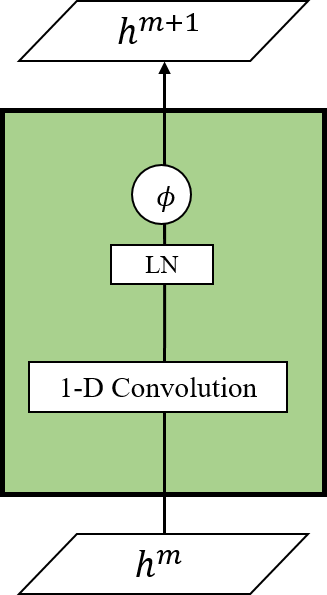}
    }
    \caption{Model description. Each box represents a single operation. $\phi$ refers to the nonlinear activation function(ReLU) and LN refers to the layer normalization\protect\cite{layernorm}. (a) The framework of DDP-GCN consists of two spatio-temporal convolutional blocks(ST-convolution blocks), and a simple 1x1 convolutional layer. (b) Each ST-convolutional block contains a spatial block and a temporal block. (c) A temporal block contains a 1-D convolution only.}
    \label{fig:model}
\end{figure*}

The system architecture of the proposed model DDP-GCN is shown in Figure \ref{fig:model}(a).
It consists of two spatio-temporal convolutional blocks(ST-convolutional blocks, Figure \ref{fig:model}(b)), and a simple 1x1 convolutional layer to reduce the number of channels at the end. 
Each ST-convolutional block contains a temporal block(Figure \ref{fig:model}(c)) and a spatial block(Figure \ref{fig:modules}, as explained in Section \ref{spatial-modules}).
We represent a variety of spatial relationships of the road network in the form of three different graph elements.
After that, we apply a simple partition filter to generate modified graph elements.
Finally, two types of multi-graph convolution are applied to effectively capture the complex spatial relationships.

\subsection{Definition of the Spatial Graph Elements}
\label{sec:spatialelements}

\begin{figure}[!t]
\centering
    \includegraphics[width=0.7\textwidth]{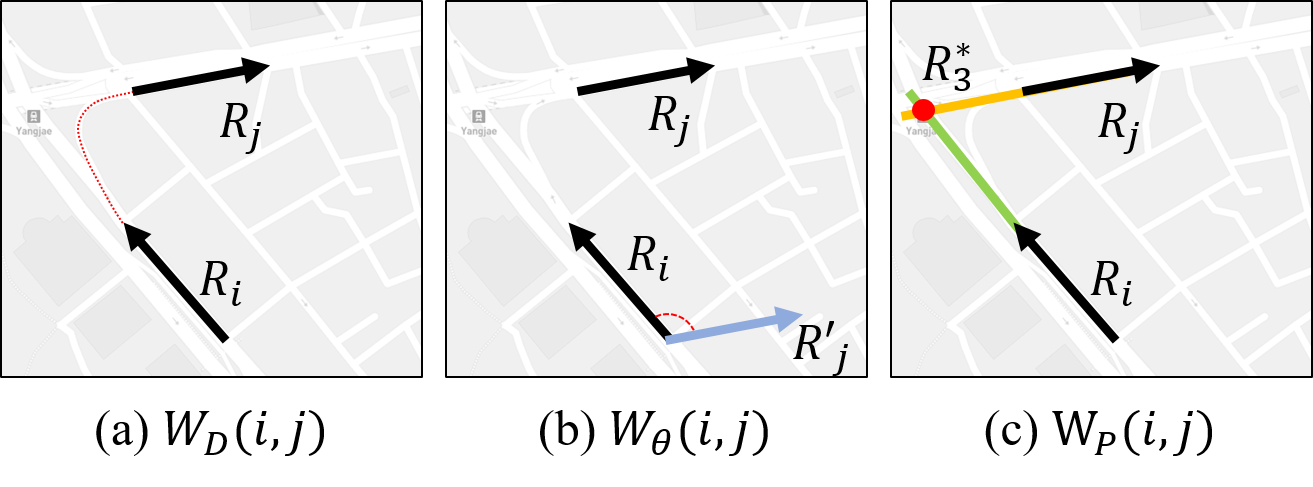}
    \vspace{-0.1cm}
    \caption{
    Three types of weighted adjacency matrices for the same link pair$(i, j)$. Each relationship is reflected into the $W(i,j)$ value.
    (a) Matrix 1: The shortest path distance ($W_D(i,j)$), (b) Matrix 2: The difference between link directions ($W_\theta(i,j)$), (c) Matrix 3: Positional relationship, two links meet at $R_3^\ast$, where the extension of $R_i$ in the forward direction meets the extension of $R_j$ in the backward direction ($W_P(i,j)$).
    }
    \label{fig:adj-matrices}
\end{figure}

Previous works \cite{chai2018bike,geng2019stmgcn,geng2019multi} defined multi-graphs on undirected networks for the bike demand forecasting. In contrast, in our work, we define multi-graphs on directed networks for traffic speed forecasting where spatial graph elements are adopted.
In order to improve our modeling capability of complex spatial relationships in road networks, 
we define two types of edge weight measures in addition to distance, as shown in Figure \ref{fig:adj-matrices}, and build proper weighted adjacency matrices 
that can be used as the spatial block graph elements.
In the following, we introduce three priors and the corresponding spatial graph elements. Only Prior 1 has been known, and the other two priors are newly introduced in our work.

\vspace{0.2cm}
\textbf{Prior 1 \cite{tobler1970}}: \textit{Everything is related to everything else. But near things are more related than distant things.}

\vspace{0.2cm}
\textbf{Graph 1 (Distance, $W_D$)}: \textit{We consider the distance $dist(i, j)$ as the shortest inter-link distance on the path. 
When the link vectors are directly connected, i.e. $R^{end}_i$ = $R^{start}_j$,
we calculate the distance as the average of the link lengths, i.e. $\frac{\left \| R_i \right \|_2 + \left \| R_j \right \|_2}{2}$.
When the link vectors are not directly connected, $R^{end}_i$ $\neq$ $R^{start}_j$,
the distance is evaluated based on the Dijkstra algorithm\cite{dijkstra1959note}.
After all the pair-wise distances are evaluated, 
we define $W_D$ using the thresholded Gaussian kernel\cite{shuman2013graphsignal} as below, where 
$\sigma$ and $\kappa$ are hyperparameters.}
\begin{equation}
    W_D(i,j) = 
    \begin{cases}
    e^{(-dist(i, j)^2 / \sigma^2)} & 
    i \neq j \text{ and } \\& 
    e^{(-dist(i, j)^2 / \sigma^2)} \geq \kappa
    \\
    0 & \text{otherwise}
    \end{cases}
\label{eq:Wdistance}
\end{equation}

Theoretically speaking, deep neural network can model any function according to universal approximation theorem\cite{cybenko1992approximation}. Therefore, one can argue that there must be a deep neural network that can achieve a high performance even when $W_D$ is modeled in a functional form different from Eq. (\ref{eq:Wdistance}). Nonetheless, the functional form in Eq. (\ref{eq:Wdistance}) has become the standard for modeling pair-wise distance~\cite{yu2017stgcn,lin2018predicting,li2017dcrnn,cui2018reachability} because it allows learning of the deep neural network parameters easier. Also, the thresholded Gaussian kernel weighting function shown in Eq. (\ref{eq:Wdistance}) is known as a common approach for defining the weight of an edge connecting vertices $i$ and $j$ when the edge weights are not naturally defined by an application~\cite{shuman2013graphsignal}.

\vspace{0.2cm}
\textbf{Prior 2}: \textit{Distant links can be related depending on their directions. 
Links having the same directions might be more related than links having the opposite directions.}

\vspace{0.2cm}
\textbf{Graph 2 (Direction, $W_\theta$)}: 
\textit{We consider a simple direction measure with a proper normalization.
To our best knowledge, we are the first to utilize the relative direction information with graph convolutional networks.}
\begin{equation}
    W_\theta(i, j)=\frac{\mod{\left (\angle{R_i}-\angle{R_j}, 2\pi\ \right )}}{2\pi}
    \label{eq:Wdirection}
\end{equation}

Obviously, Prior 2 might be applicable to only a small portion of the distant link pairs.  
By providing direction information through $W_\theta$ into GCN, however, we are making it much easier for the deep neural network to learn direction based patterns that are related to Prior 2, if any, and to utilize them for prediction. If the dataset of interest does not have any strong pattern related to Prior 2, the model will simply not perform any better with $W_\theta$.
Prior 2 might be also helpful for modeling local connections. For instance, we can think of two consecutively connected links. If both have similar link directions, their role as a whole would be different from the case where the two links have 90 degree difference in their link directions.

\vspace{0.2cm}
\textbf{Prior 3}: \textit{Links are related depending on how they can be connected. The positional relationship, such as whether two links are heading to closely located destinations, could be additionally informative to the distance for understanding how links interact.}

\vspace{0.2cm}
\textbf{Graph 3 (Positional relationship, $W_P$)}: 
\textit{While Graph 1 considers only the shortest path-distance that connects from $R_i^{end}$ to $R_j^{start}$, two links can be connected through many different paths.
In order to capture how two links are related while considering a variety of connection paths, 
we extend the link vectors maintaining the start point $R_i^{start}$ and the link direction $\angle{R_i}$.
Then we define $W_P=\left\{ W_P^1, W_P^2, W_P^3, W_P^4 \right\}$ as the unweighted adjacency matrices that contain the information on where the two extended link vectors meet as described below for each $k \in \left\{ 1, 2, 3, 4\right\}$.}
\begin{equation}
    W_P^k(i, j) = 
    \begin{cases}
    1, R^\ast_k \text{ exists} \\
    0, \text{otherwise}
    \end{cases}
    \label{eq:positional-relationship}
\end{equation}

\begin{figure}[t!]
\centering
    \includegraphics[width=0.8\textwidth]{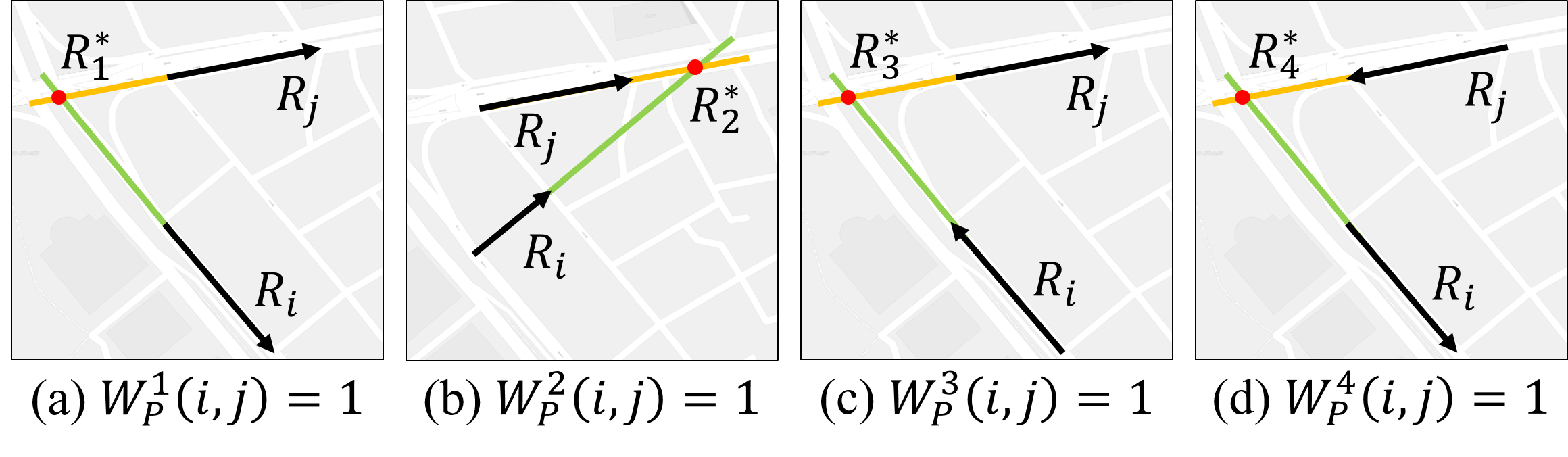}
    \caption{The four positional relationships. Each represents a different type of $W_P$, depending on where the two extended link vectors meet.
    }
    \label{fig:cdiop-def}
\end{figure}

Figure \ref{fig:cdiop-def} shows the four types of positional relationships.
$R_k^\ast$ represents four possible intersection points of two extended links vectors. $R_1^\ast$ exists if the extension of $R_i$ in the backward direction meets the extension of $R_j$ in the backward direction. $R_2^\ast$, $R_3^\ast$, and $R_4^\ast$ are similarly defined depending on whether forward or backward extensions of $R_i$ and $R_j$ can meet.
Each type implies how two links could interact. For example, two links would head to the same area when $R_2^\ast$ exists as described in Figure \ref{fig:cdiop-def}(b). 
Note that any $R_k^\ast$ cannot exist when two links are exactly parallel. This case is taken into consideration by setting all $W_P^k$ values to zero.
In our experiment, we consider the dataset of Seoul where some of the business districts can become major destination locations or some of the commuter towns can become major departure locations. For such a crowded city, $W_P$ might be able to play an essential role, especially when the model is flexible enough to extract the complicated positional relationship patterns in a useful way. {For the highly complex urban network of Seoul, often there are multiple major routes that connect from or to a commuter town or a business district, and the routes are likely to have a strong correlation with the positional relationship.
Indeed, the positional relationship turned out to be useful for our dataset as will be shown in Section~\mbox{\ref{subsec:ablation_study}}. In general, however, the new spatial relationship has not been studied before and there is a risk that its effectiveness might not transfer well to other traffic environments. Similar to our positional relationship, several new types of spatial relationships have been studied in recent years. Examples include speed limit\cite{shin2020incorporating}}, functional similarity\cite{yao2018dmvst}, and transportation connectivity\cite{geng2019stmgcn}. With the increasing data size, some of these new endeavors might turn out to be generally effective.

\subsection{Partition Filters}

\begin{figure}[h!]
\centering
    \includegraphics[width=0.75\textwidth]{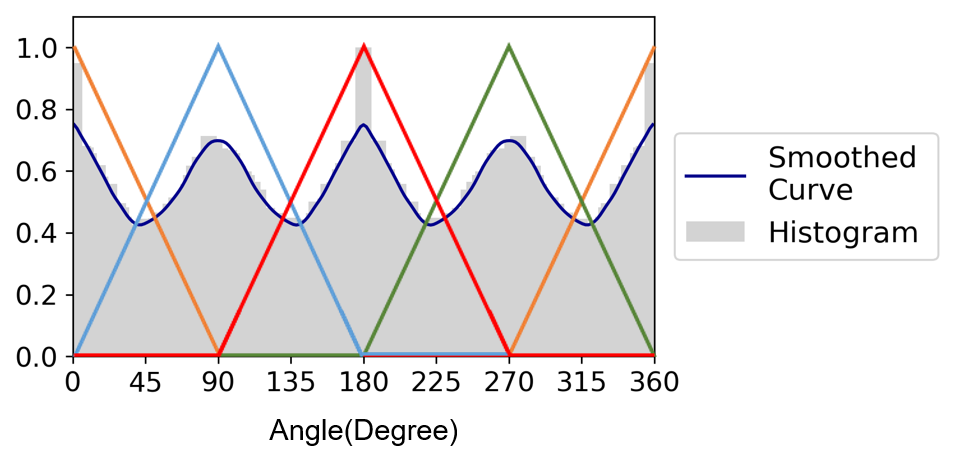}
    \caption{Creating partition filters from $W_\theta$. 
    X axis shows angles and Y axis shows $t_m(W_\theta)$ values for the choice of partition filters.
    The direction histogram of the links is properly scaled for plotting purpose and smoothed with a Gaussian kernel. In our work, the four directions $\{0, 90, 180, 270\}$ degrees, happened to coincide with the histogram peaks, however, this is not the only possible choice, and the design of the partition filter allows any number and any choice of directions based on the histogram analysis.
    }
    \label{fig:partition-filters}
\end{figure}

Urban road networks are often based on grid structure where most of the roads head into one of the four directions. Rather than hoping deep learning models to learn such a grid concept through data, we propose a method to indicate the four directions explicitly into the spatial graph elements.
For this purpose, we define a set of partition filters $\{t_m\}$ that can be used over $W$ to create 
$t_m(W)=W_m\in\left \{{W}_{1}, \dots, {W}_{M} \right \}$. 
$t_m(.)$ is a scalar-input scalar-output function, and $t_m(W)$ describes the element-wise application of filter $t_m(.)$. We constrain the set of partition filters $T_M=\{t_1, ..., t_M\}$ to satisfy $\sum_{m=1}^M{{t}_m(W)} = W$ in order to make sure that $W$ is spread over the $M$ partitioned matrices without any increase or decrease in the element-wise sum. 
$M$ can be designed based on the histogram analysis.
To smoothly handle the boundary values, we choose triangular partition filters.

In our work, only $W \in \left \{W_\theta, W_D\right \}$ are considered for applying the partition filters.
When four partition filters are applied to $W_\theta$ with appropriate filter shapes, we can help deep neural network learn the concept of grid in the dataset. Figure \ref{fig:partition-filters} shows 4-directional triangular partition filters, $T_{M=4}$, that we have used in our experiments. We have also applied partition filters to $W_D$.
For $W_D$, appropriate partition filters can be designed by investigating the density in a similar way as in  Figure \ref{fig:partition-filters}. In this case, the partition filters are used to group link pairs according to their distance range.
We note that the partition filter has an analogy to the edge-conditioned filters 
introduced in \cite{simonovsky2017dynamic} for convolutional neural networks on graphs. 
Our partition filters, however, are static while the filter weights in \cite{simonovsky2017dynamic} are dynamically generated for each specific input sample. 

\subsection{Building a Spatial Block}
\label{spatial-modules}

\begin{figure*}[h]
\centering
    \subfloat[Single]{
        \includegraphics[height=16em]{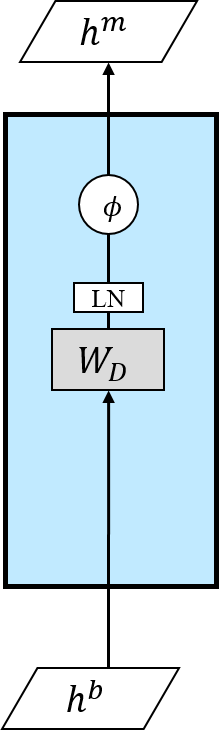}
    } \rulesep
    \subfloat[Parallel]{
        \includegraphics[height=16em]{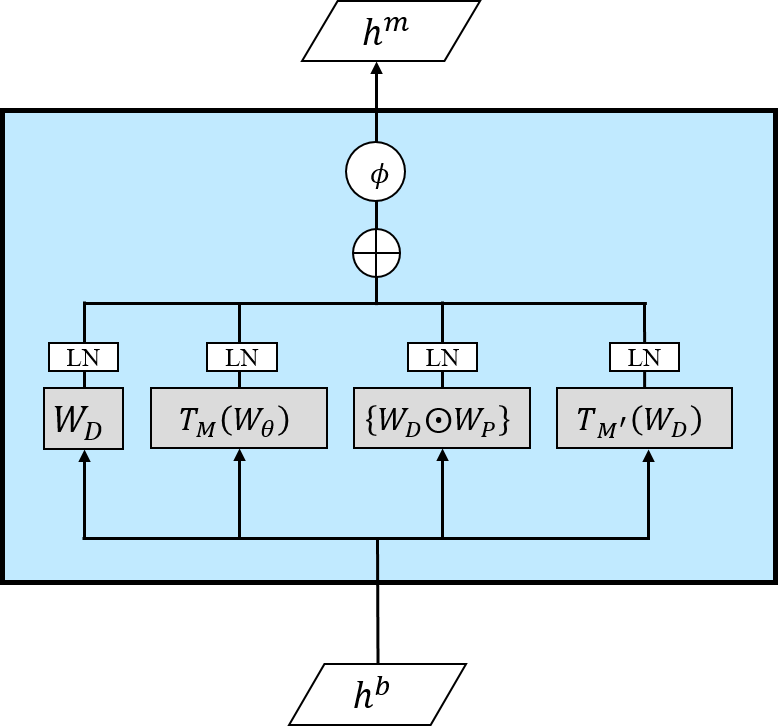}
    } \rulesep
    \subfloat[Stacked]{
        \includegraphics[height=16em]{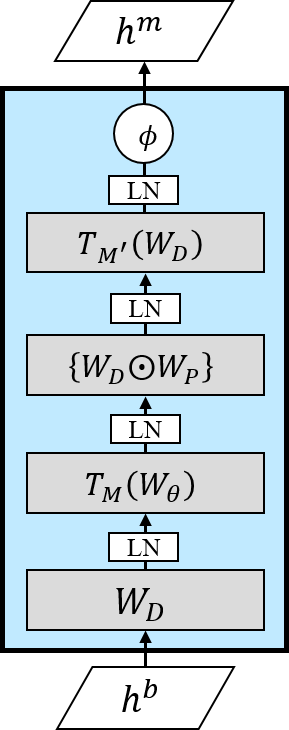}
    }
    \hspace{0.3cm}
    \caption{Three types of spatial blocks. Each small gray box represents a single graph convolution element and we only denote the weighted adjacency matrix for simplicity. $\oplus$ refers to the element-wise sum. The box with LN indicates layer normalization. Details are provided in the text.
    }
    \label{fig:modules}
\end{figure*}

With the newly defined graph elements, we are ready to build spatial blocks for extracting complex spatial relationships. 
Spatial blocks are based on the three graph elements $W_D, W_\theta$, and $W_P$, and their partitioned versions. 
We use $\left\{ W_D \odot W_P \right\}$ instead of $W_P$ to exploit the hybrid information.

In Figure \ref{fig:modules}, each gray box represents a single graph convolution as described in Eq. (\ref{def:gc}), and we only denote the weighted adjacency matrix in the figure for simplicity.
The number of convolutional operation and the choice of weight matrices can vary depending on the dataset or the prediction task. 

We designed three types of spatial blocks, as shown in Figure \ref{fig:modules}.
\textit{Single}(Figure \ref{fig:modules}(a)) refers to the simple graph convolution considering only the Euclidean distance.
\textit{Parallel}(Figure \ref{fig:modules}(b)) and \textit{Stacked}(Figure \ref{fig:modules}(c))
refer to the multi-graph convolution including distance, direction, and positional relationship information.
While both utilize four convolutional operations with different graph elements individually, they have different structures for connecting the graph elements. \textit{Parallel} structure can be regarded as equivalent to multi-graph convolution structure as defined in previous works \cite{chai2018bike,geng2019stmgcn,geng2019multi}.

\subsection{Building a Temporal Block}
We conducted extensive experiments to design the temporal blocks, including graph convolution, self-attention\cite{vaswani2017attention}, multi-convolution\cite{szegedy2017inception}, and temporal relational reasoning\cite{zhou2018temporalrelation}. But it turns out that simple convolution showed the best performance with the shortest training time. Based on the empirical results, we adapt the simple temporal block as shown in Figure \ref{fig:model}(c), where we choose the kernel size of 3 and stride of 1.
\section{Experiment}
\label{experiment}
In this section, experiment settings and results are presented. We provide ablation test results as well\footnote{The codes are available at \url{https://github.com/snu-adsl/DDP-GCN}.}.

\subsection{Datasets}

We conducted experiments on two real-world large-scale datasets of Seoul, South Korea\footnote{The original source of dataset is http://topis.seoul.go.kr/, and the processed dataset that was used for this study is available at https://github.com/snu-adsl/ddpgcn-dataset with its descriptions in English.}.
Urban1 (Gangnam) and Urban2 (Mapo) correspond to the most crowded regions in Seoul, and they have highly complex connectivity patterns where
most of the streets have bidirectional links with complicated traffic signals. The traffic data were collected for a month ranging from Apr 1st, 2018 to Apr 30th, 2018. The datasets were collected using GPS of over 70,000 taxis, where the trajectory samples were collected every 5 minutes and the post data processing was applied to calculate the average traffic speed of each link.

\begin{table}[h!]
\centering
\caption{\label{table:dataset}Details of the datasets.}
\resizebox{0.7\textwidth}{!}{%
\begin{tabular}{c||c|c}
\hline
Dataset                                                                              & Urban1         & Urban2        \\ \hline
Time spans                                                               & \multicolumn{2}{c}{4/1/2018 $\sim$ 4/30/2018}          \\ \hline
Time interval                                                            & \multicolumn{2}{c}{5min}                        \\ \hline
Region size (width, height)(m)  & \multicolumn{2}{c}{(7000, 7000)}    \\ \hline
Number of links                                                                  & 480            & 455           \\ \hline
Speed mean(std)(km/h)         & 26.333(10.638) & 25.917(9.784) \\ \hline
Length mean(min, max)(m)                                              & 592(171, 2622) & 561(80, 2629) \\ \hline
Links per $\text{km}^2$ & 11.274 & 10.280 \\ \hline
\end{tabular}
}
\label{table:dataset}
\end{table}

\begin{figure}
    \centering
    \subfloat[]{
    \includegraphics[height=0.28\textheight]{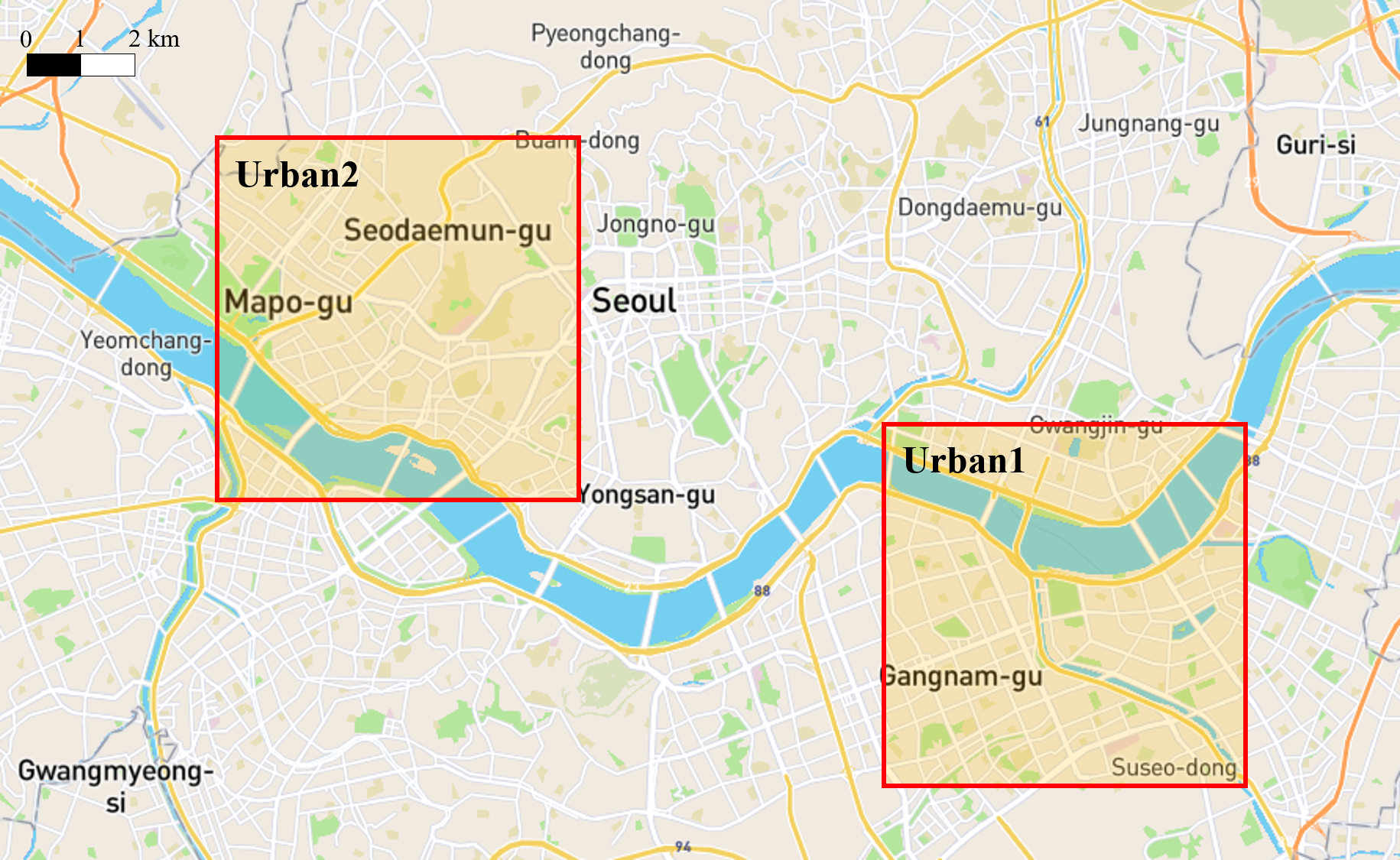}
    } \hspace{0.1cm}
    \subfloat[]{
    \includegraphics[height=0.28\textheight]{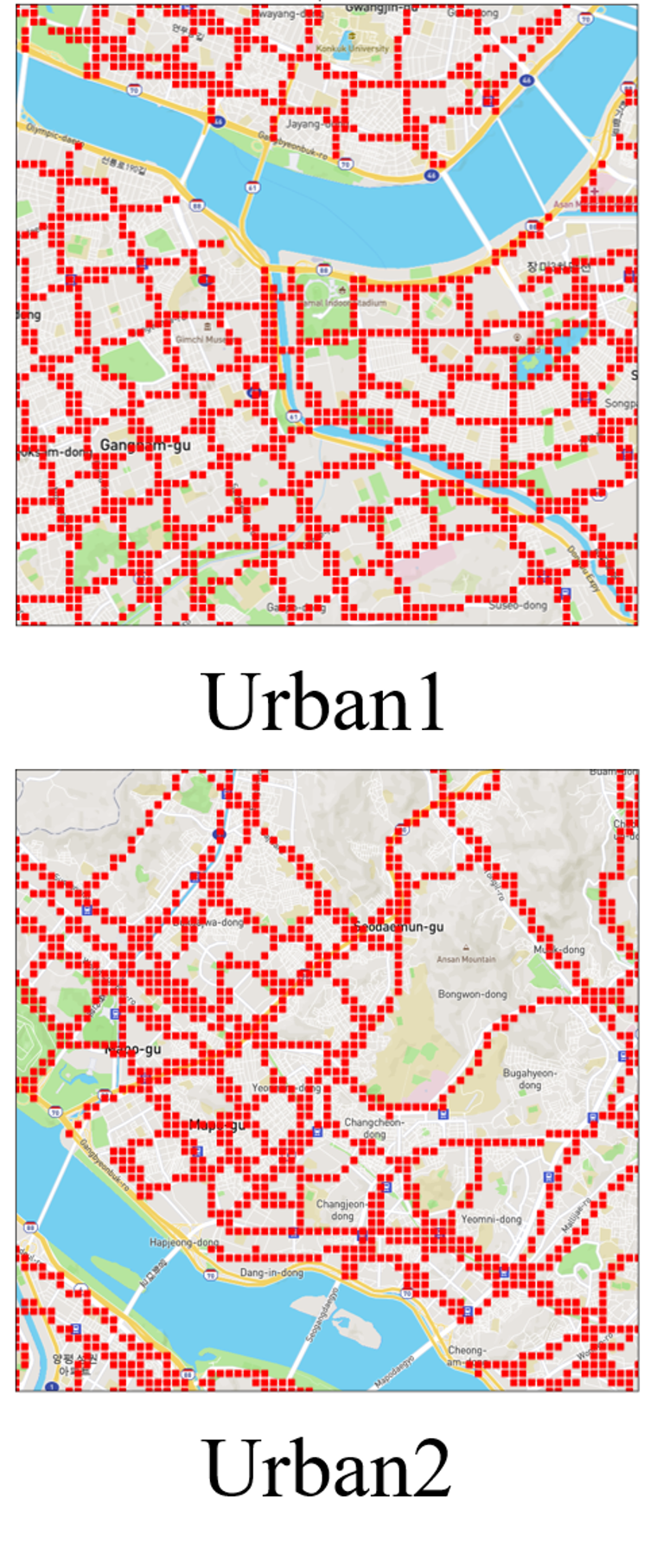}
    }
    \caption{Geographical coverage of Urban1(Gangnam) and Urban2(Mapo) in the city of Seoul. (a) Map of Seoul and the two urban areas. (b) The road links under consideration are marked with red dots.}
    \label{fig:dataset}
\end{figure}

Most of the previous studies on traffic forecasting have focused on traffic networks with  freeways only and defined links as simple points without direction information \cite{li2017dcrnn,yu2017stgcn,yu2019stunet,chen2019gated,cui2018reachability}.
Our datasets, however, contain complex urban networks with a large number of intersections and traffic signals. Also, the links are defined as vectors with direction information.
A summary of Urban1 and Urban2 datasets are provided in Table \ref{table:dataset}, and geographical coverage is shown in Figure \ref{fig:dataset}.
While not shown in the table, another characteristics of our datasets is reachability. For each dataset, every single link is reachable from any other link in the dataset. In contrast, only 27$\%$ of link pairs are reachable to each other in the METR-LA dataset studied in \cite{li2017dcrnn,yu2019stunet}.

\subsection{Experimental Settings}
We applied Z-score normalization for both datasets. After excluding the weekends, 70\% of the data was used for training, 10\% for validation, and the remaining 20\% for testing. For $W_D$, as defined in Eq. (\ref{eq:Wdistance}),
$\sigma$ and $\kappa$ are set depending on the data scale. In our study, we set $\sigma$ to be $10^6$ and $\kappa$ to be 0.
For partition filters, we determined the number of filters, $M$ for $W_\theta$ and $M'$ for $W_D$, based on the histogram analysis of each dataset as described in Section \ref{sec:spatialelements}.
We set $M$ to be 4 (Urban1, Urban2) and $M'$ to be 3 (Urban1) and 4 (Urban2).
We set both $T'$ and $T$ as 12 samples, where 12 corresponds to an one hour span.
All experiments were implemented using Tensorflow 1.15 on a Linux cluster(CPU: Intel(R) Xeon(R) CPU E6-2620 v4 @ 2.10GHz, GPU: NVIDIA TITAN V). 

We compared the proposed model (DDP-GCN) with the following baslines:
(1) HA: Historical Average; 
(2) VAR: Vector Auto-Regression\cite{hamilton1994time}; 
(3) LSVR: Linear Support Vector Regression; (4) ARIMA: Auto-Regressive Integrated Moving Average model;
(5) FC-LSTM: Recurrent Neural Network with fully connected LSTM hidden units\cite{sutskever2014sequence}; (6) DCRNN: Diffusion Convolutional Recurrent Neural Networks\cite{li2017dcrnn}, which manipulates bidirectional diffusion convolution on the graph for capturing spatial dependency and uses sequence-to-sequence architecture with gated recurrent units to capture temporal dependency, and (7) STGCN: Spatio-Temporal Graph Convolutional Networks\cite{yu2017stgcn}, which is composed of spatiotemporal convolutional blocks including two gated sequential convolution layers and one spatial graph convolution layer in between.

\subsection{Performance Comparison}

\begin{table*}[t!]
\centering
\caption{Performance comparison for (a) Urban1 dataset and (b) Urban2 dataset. The best performance in each category is indicated in \textbf{bold}.
}
    \resizebox{\textwidth}{!}{%
        \begin{tabular}{c||ccc|ccc|ccc}
        \multicolumn{10}{@{}c}{(a) Urban1} \\
        \hline
        \multirow{2}{*}{Algorithm} & \multicolumn{3}{c|}{30min} & \multicolumn{3}{c|}{45min} & \multicolumn{3}{c}{60min} \\ \cline{2-10}
                      & MAPE(\%)    & MAE    & RMSE   & MAPE(\%)    & MAE    & RMSE   & MAPE(\%)    & MAE    & RMSE   \\ \hline\hline
        HA                  & 14.68   & 3.34   & 5.42   & 14.67   & 3.34   & 5.42   & 14.68   & 3.34   & 5.41   \\ \hline
        VAR                 & 23.10   & 5.06   & 7.04   & 22.82   & 4.99   & 6.92   & 22.73   & 4.97   & 6.88   \\ \hline
        LSVR                & 15.35   & 3.82   & 5.64   & 17.99   & 3.89   & 5.74   & 17.39   & 3.93   & 5.84   \\ \hline
        ARIMA               & 15.40   & 3.49   & 5.28   & 16.85   & 3.79   & 5.65   & 18.09   & 4.04   & 5.94   \\ \hline
        FC-LSTM             & 17.29   & 3.91   & 6.38   & 17.32   & 3.92   & 6.39   & 17.31   & 3.92   & 6.39   \\ \hline
        DCRNN               & \textbf{13.52}   & 3.17   & 4.94   & 14.83   & 3.46   & 5.30   & 15.95   & 3.73   & 5.61   \\ \hline
        STGCN               & 14.38   & 3.07   & 4.57   & 16.72   & 3.42   & 4.83   & 19.37   & 3.80   & 5.04   \\ \hline\hline
        DDP-GCN(Single)     & 13.83   & 3.06   & 4.54   & 13.82   & 3.06   & 4.54   & 15.11   & 3.30   & 4.95   \\ \hline
        DDP-GCN(Parallel)   & 13.95   & 3.09   & 4.62   & 13.94   & 3.09   & 4.62   & 13.77   & 3.06   & 4.61   \\ \hline
        DDP-GCN(Stacked)    & 13.60   & \textbf{3.00}   & \textbf{4.46}   & \textbf{13.59}   & \textbf{3.00}   & \textbf{4.46}   & \textbf{13.53}   & \textbf{2.99}   & \textbf{4.47}   \\ \hline
        \end{tabular}
    }
\\ \bigskip
\centering
    \resizebox{\textwidth}{!}{%
        \begin{tabular}{c||ccc|ccc|ccc}
        \multicolumn{10}{@{}c}{(b) Urban2} \\
        \hline
        \multirow{2}{*}{Algorithm} & \multicolumn{3}{c|}{30min} & \multicolumn{3}{c|}{45min} & \multicolumn{3}{c}{60min} \\ \cline{2-10}
                      & MAPE(\%)    & MAE    & RMSE   & MAPE(\%)    & MAE    & RMSE   & MAPE(\%)    & MAE    & RMSE   \\ \hline\hline
        HA                  & 14.43   & 3.23   & 4.86   & 14.42   & 3.22   & 4.86   & 14.41   & 3.22   & 4.85   \\ \hline
        VAR                 & 20.82   & 4.58   & 6.31   & 20.55   & 4.52   & 6.22   & 20.43   & 4.49   & 6.19   \\ \hline
        LSVR                & 17.01   & 4.38   & 5.83   & 16.82   & 4.22   & 5.71   & 18.45   & 3.92   & 5.37   \\ \hline
        ARIMA               & 14.78   & 3.30   & 4.77   & 15.99   & 3.56   & 5.09   & 17.03   & 3.78   & 5.37   \\ \hline
        FC-LSTM             & 17.10   & 3.81   & 5.57   & 17.12   & 3.81   & 5.58   & 17.12   & 3.82   & 5.58   \\ \hline
        DCRNN               & 13.55   & 3.08   & 4.58   & 14.63   & 3.31   & 4.86   & 15.52   & 3.50   & 5.08   \\ \hline
        STGCN               & 14.02   & 2.99   & 4.37   & 15.82   & 3.33   & 4.79   & 17.78   & 3.69   & 5.26   \\ \hline\hline
        DDP-GCN(Single)     & 13.34   & 2.93   & 4.26   & 13.33   & 2.93   & 4.26   & 13.99   & 3.06   & 4.43   \\ \hline
        DDP-GCN(Parallel)   & 13.37   & 2.94   & 4.29   & 13.35   & 2.94   & 4.29   & 13.21   & 2.91   & 4.25   \\ \hline
        DDP-GCN(Stacked)    & \textbf{13.12}   & \textbf{2.88}   & \textbf{4.21}   & \textbf{13.11}   & \textbf{2.88}   & \textbf{4.21}   & \textbf{13.09}   & \textbf{2.87}   & \textbf{4.21}   \\ \hline
        \end{tabular}
    }
    \label{table:performance}
    \end{table*}

Table \ref{table:performance} shows the performance results of DDP-GCN and the baseline models. 
Each model predicted all 12 sequential traffic speed values of all the links simultaneously, and the accuracy of 30 minute, 45 minute, and 60 minute predictions are summarized in Table \ref{table:performance}.
Three commonly used metrics in traffic forecasting were evaluated, including
(1) Mean Absolute Error (MAE), (2) Mean Absolute Percentage Error (MAPE), and (3) Root Mean Squared Error (RMSE).
We have repeated each experiment five times, and the average performance is reported in Table \ref{table:performance}. 
For all the experiments and all three metrics, the standard deviation was very small (less than 0.05 except for STGCN that had up to 0.2 of standard deviation).

One of our proposed models, DDP-GCN(Stacked), achieved the best performance for all cases except for the case of 30 minute MAPE of Urban1. For 60 minute forecasting, our model showed an improvement of 7.52\% on average (10.14\% maximum) when compared to the previous state-of-the-art performance. Among our three models, the stacked spatial block, firstly introduced in our work, outperformed the parallel spatial block for all cases.
While the previous state-of-the-art GCN methods, i.e. DCRNN and STGCN, suffer when 
forecasting horizon is increased, DDP-GCN(Stacked)'s performance is sustained even when the prediction time is increased from 30 minutes to 60 minutes. 

Interestingly, we can observe that HA tends to be reasonably accurate even though the method is extremely simple. 
We believe that this result is mainly due to the strong weekly periodicity of our dataset. While the other methods utilize only the last 1-hour data to predict the future 1-hour period, HA utilizes a different type of information. It looks up the speed information of the same day and same time of the previous weeks. 
While our DDP-GCN models exploit only the last 1-hour information, our model almost always outperformed the other methods including HA. It suggests that it might be possible to further improve DDP-GCN's performance by utilizing weekly data in addition to the last 1-hour data.
Also, it suggests that the non-Euclidean spatial relationships, direction, and positional relationship, are quite powerful priors for the accurate speed forecasting.

When DDP-GCN(Stacked) is compared to the other two graph convolution based methods only (the better performing one between DCRNN and STGCN for each category), DDP-GCN(Stacked) showed 9.75\% average (19.84\% maximum) improvement over all forecasting horizons and 16.18\% average improvement for 60 minute forecasting horizon. While the improvements were large, we note that DDP-GCN tends to provide enhancements for the longer forecasting horizons. For shorter forecasting horizons such as 15 minute, DDP-GCN actually performed worse than DCRNN and STGCN. A possible explanation is that information in time dimension is sufficient for each link's short-term forecasting while information in space dimension is essential for long-term forecasting.

\begin{figure*}[th!]
\centering
    \subfloat[]{
        \includegraphics[width=0.45\textwidth]{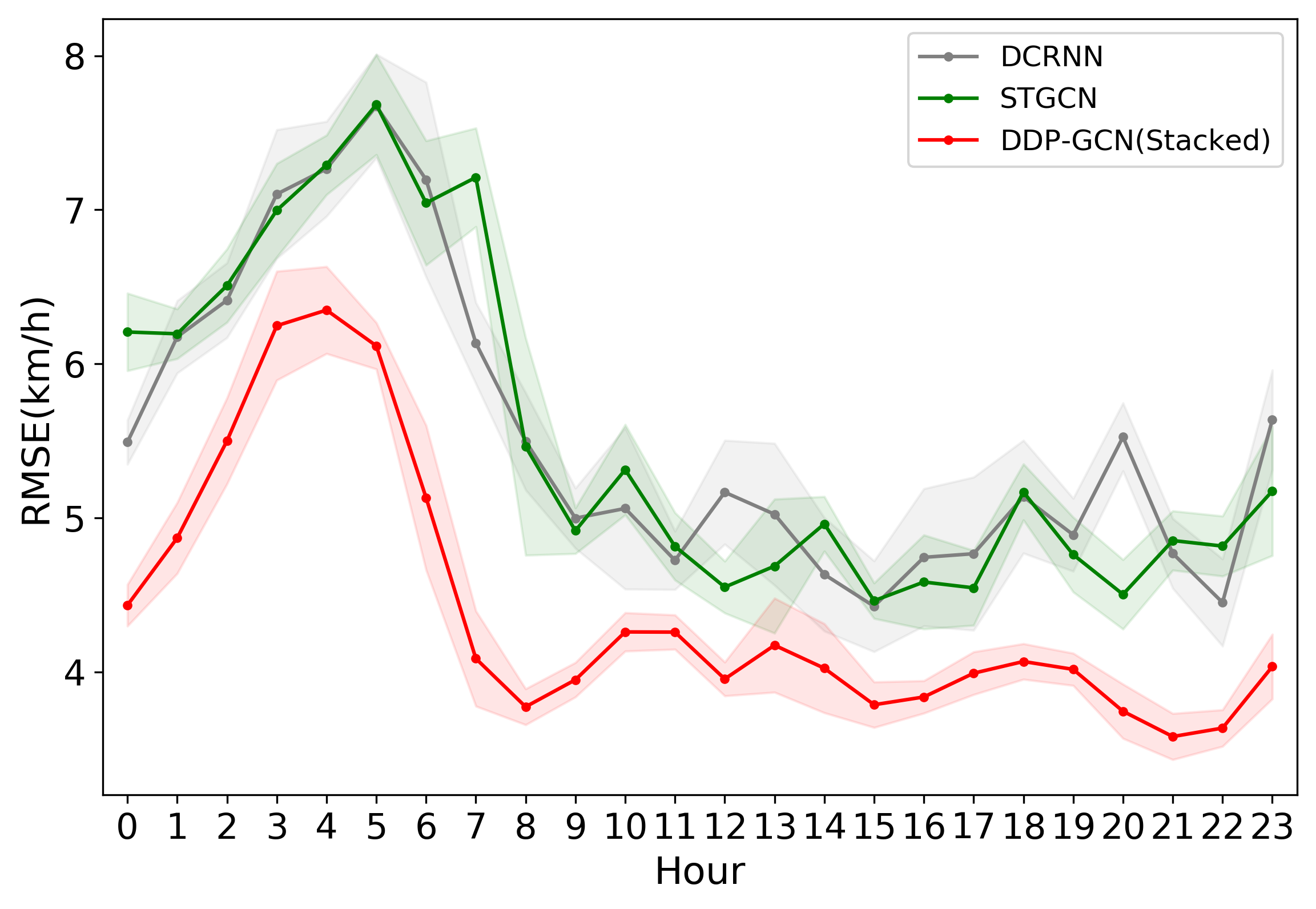}
        } \hspace{0.2cm}
    \subfloat[]{
        \includegraphics[width=0.47\textwidth]{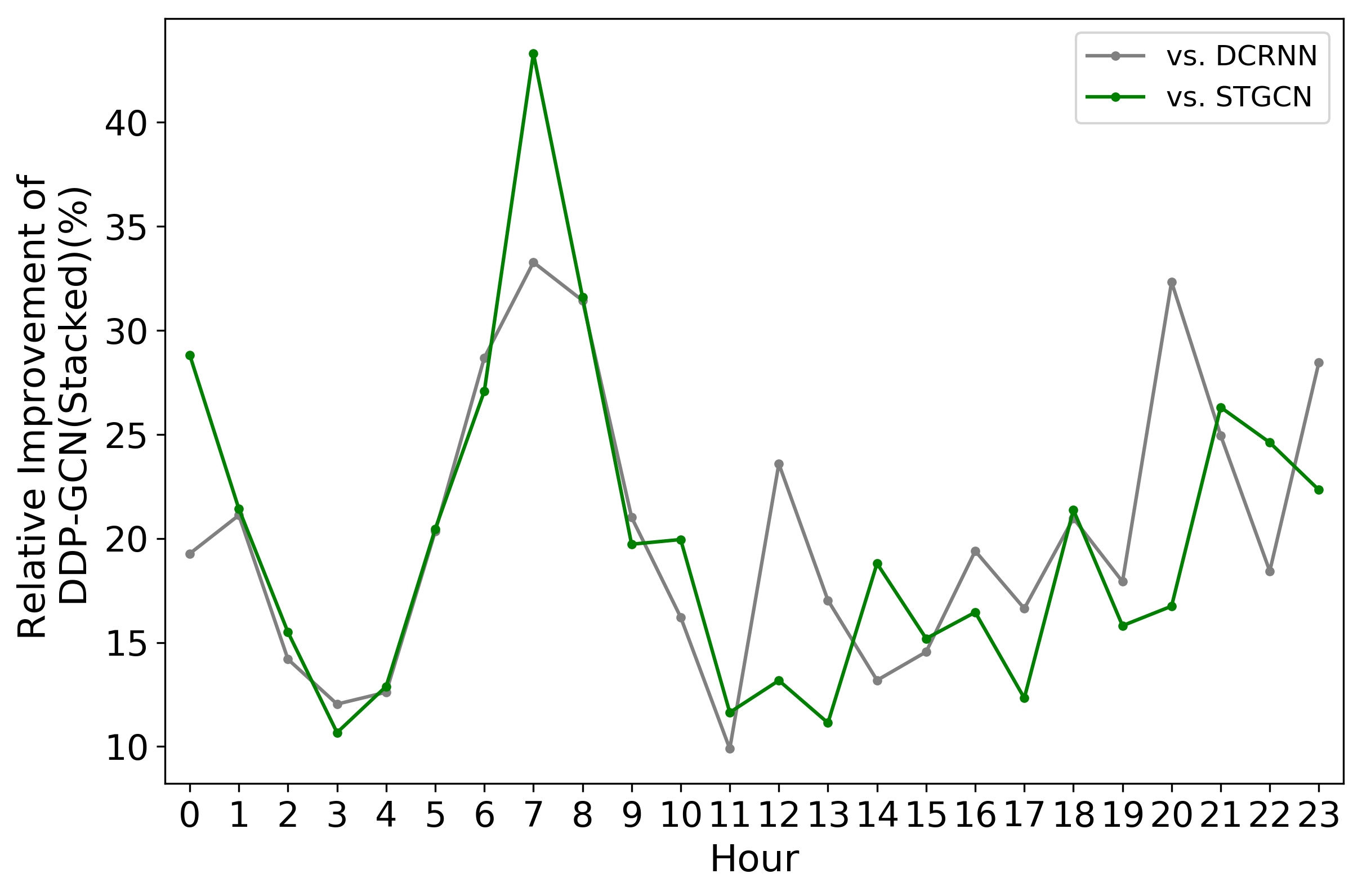}
        }
    \caption{Hourly performance of DCRNN, STGCN, and DDP-GCN(Stacked) for Urban1 60min. (a) RMSE performance curves where shading indicates confidence interval of one standard deviation. (b) Relative improvements over DCRNN and STGCN. The improvements can be as large as 20$\sim$45\% during the morning commute hours (6$\sim$9am).}
    \label{fig:hourly-analysis}
\end{figure*}

Hourly analysis results are shown in Figure \ref{fig:hourly-analysis}(a) where the shape of RMSE curves follows the shape of average traffic speed. Even though not shown here, the shape of average traffic speed looks very similar to the shape of the shown RMSE curves. Because there is almost no traffic in the early morning, the average traffic speed is high. In consequence, the high speed results in a large RMSE simply because of the scaling effect. In (b), relative RMSE improvements are shown against DCRNN and STGCN. It is interesting to note that DDP-GCN's relative improvement is clearly larger during the morning commute hours. The result indicates that our spatial graph elements are most valuable when the traffic dynamics are most complicated.

\begin{figure*}[t!]
\centering
    \includegraphics[width=0.97\textwidth]{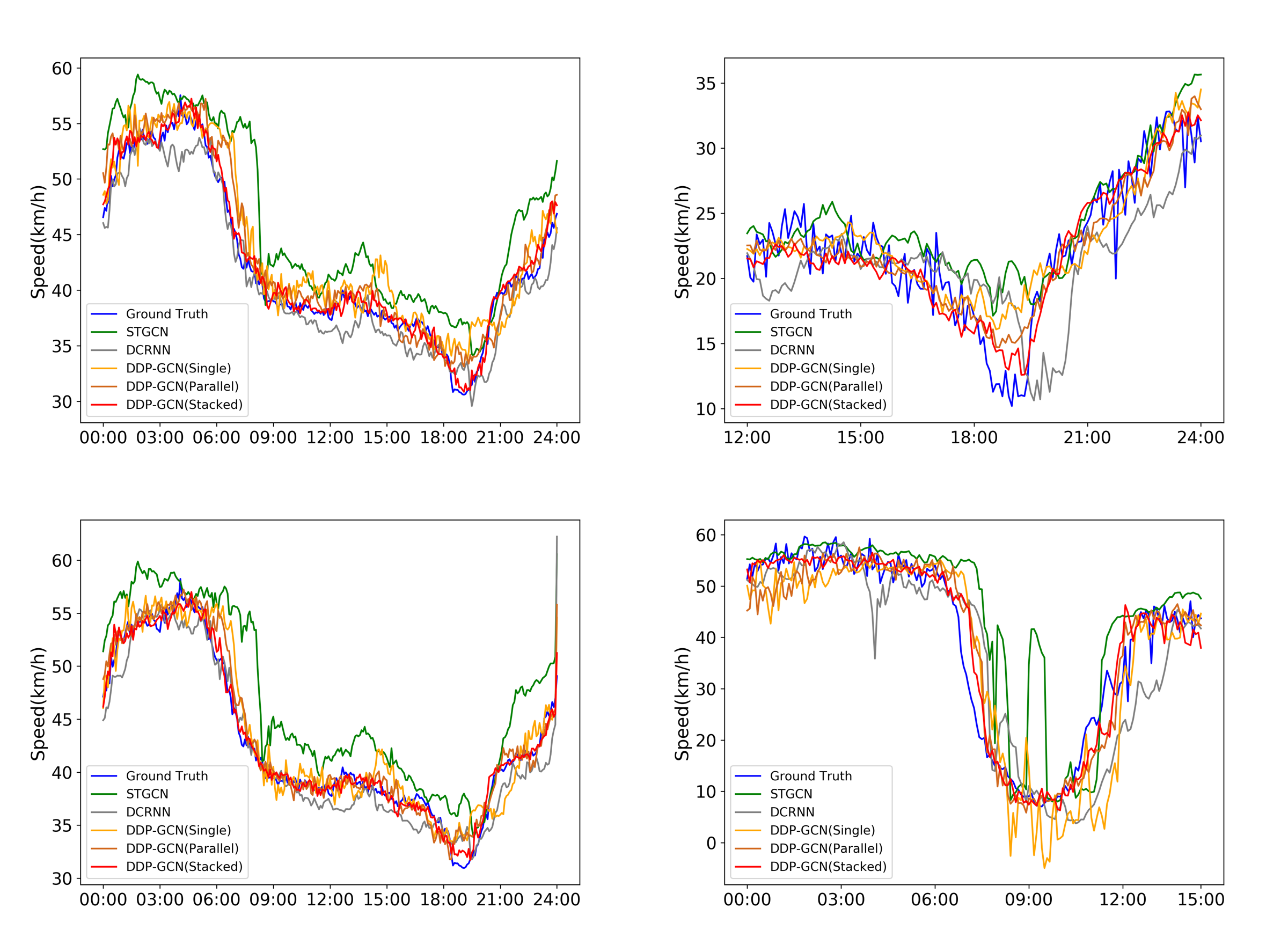}
    \caption{
    Examples of 60 minute prediction results for the dataset Urban1. (Upper Left: Apr 25th, Upper Right: Apr 26th, Lower Left: Apr 27th, Lower Right: Apr 30th) DDP-GCN can smoothly track the ground truth.}
    \label{fig:prediction-results}
\end{figure*}

Segments of the prediction results are shown in Figure \ref{fig:prediction-results}. We can easily see that DDP-GCN(Stacked) is superior at tracking the ground truth curves. This result suggests that non-Euclidean information is essential for capturing abrupt changes in complex networks in advance.

\subsection{Ablation Test of Spatial Graph Elements and Partition Filters}
\label{subsec:ablation_study}

Each spatial graph element is illustrated as a small gray box in Figure \ref{fig:modules}, and it 
contains one graph convolutional operation. To investigate the contribution of each spatial graph element, we evaluated the performance degradation through ablation test.

\begin{table*}[h!]
\vspace{0.3cm}
\caption{Ablation and replacement test results. The case with the most performance degradation is indicated in \textbf{bold}.}
\centering
\resizebox{\textwidth}{!}{%
\subfloat[Removing element]{
    \begin{tabular}{c||c|c|c}
    \hline
    Removed Element & MAPE(\%) & MAE    & RMSE \\ \hline\hline
    Distance  & 13.52 & 3.00 & 4.48 \\ \hline
    Direction & 13.66 & 3.02  & 4.54 \\ \hline
    \textbf{Positional Relationship} & \textbf{13.71} & \textbf{3.03} & \textbf{4.55} \\ \hline
    Distance($T_M'(W_D)$)       & 13.62 & 3.02 & 4.53 \\ \hline\hline
    None    & 13.53 & 2.99 & 4.47 \\ \hline
    \end{tabular}
    }
    \hspace{0.8cm}
\subfloat[Element replaced with $W_D$]{
    \begin{tabular}{c||c|c|c}
    \hline
    Replaced Element & MAPE(\%) & MAE    & RMSE \\ \hline\hline
    Distance (no change)    & 13.53 & 2.99 & 4.47 \\ \hline
    Direction & 13.66 & 3.02  & 4.52 \\ \hline
    \textbf{Positional Relationship} & \textbf{13.72} & \textbf{3.03} & \textbf{4.55} \\ \hline
    Distance ($T_M'(W_D)$)       & 13.60 & 3.01 & 4.52 \\ \hline
    \hline
    None    & 13.53 & 2.99 & 4.47 \\ \hline
    \end{tabular}
    }
    \vspace{-0.1cm}
    }
    \label{table:ablation}
\end{table*}

\begin{table*}[h!]
\caption{Ablation test results of partition filters. Only the partition filtering was removed within the modified element. The results of Urban1 dataset are presented, and the results for Urban2 dataset were similar. The case with the most performance degradation is indicated in \textbf{bold}.}
\centering
\resizebox{\textwidth}{!}{%
\begin{tabular}{c||ccc|ccc|ccc}
\hline
\multirow{2}{*}{\begin{tabular}[c]{@{}c@{}}Modified Element\\ (Partition filter removal)\end{tabular}} & \multicolumn{3}{c|}{30min} & \multicolumn{3}{c|}{45min} & \multicolumn{3}{c}{60min} \\ \cline{2-10}
                  & MAPE(\%)    & MAE    & RMSE   & MAPE(\%)    & MAE    & RMSE   & MAPE(\%)    & MAE    & RMSE   \\ \hline\hline
Direction  & 13.72 & 3.02 & 4.51 & 13.71 & 3.02 & 4.51 & 13.70 & 3.02 & 4.54 \\ \hline
Distance  & 13.66 & 3.02 & 4.49 & 13.65 & 3.01 & 4.49 & 13.66 & 3.02 & 4.52 \\ \hline
\textbf{Both} & \textbf{13.82} & \textbf{3.04} & \textbf{4.57} & \textbf{13.81} & \textbf{3.04} & \textbf{4.56} & \textbf{13.88} & \textbf{3.06} & \textbf{4.61}  \\ \hline\hline
None & 13.60 & 3.00 & 4.46 & 13.59 & 3.00 & 4.46 & 13.53 & 2.99 & 4.47 \\ \hline
\end{tabular}
}
\vspace{-0.1cm}
\label{table:ablation-partiton}
\end{table*}

\begin{figure*}[h!]
\centering
\vspace{-0.1cm}
    \subfloat[Urban1]{
        \includegraphics[width=0.5\textwidth]{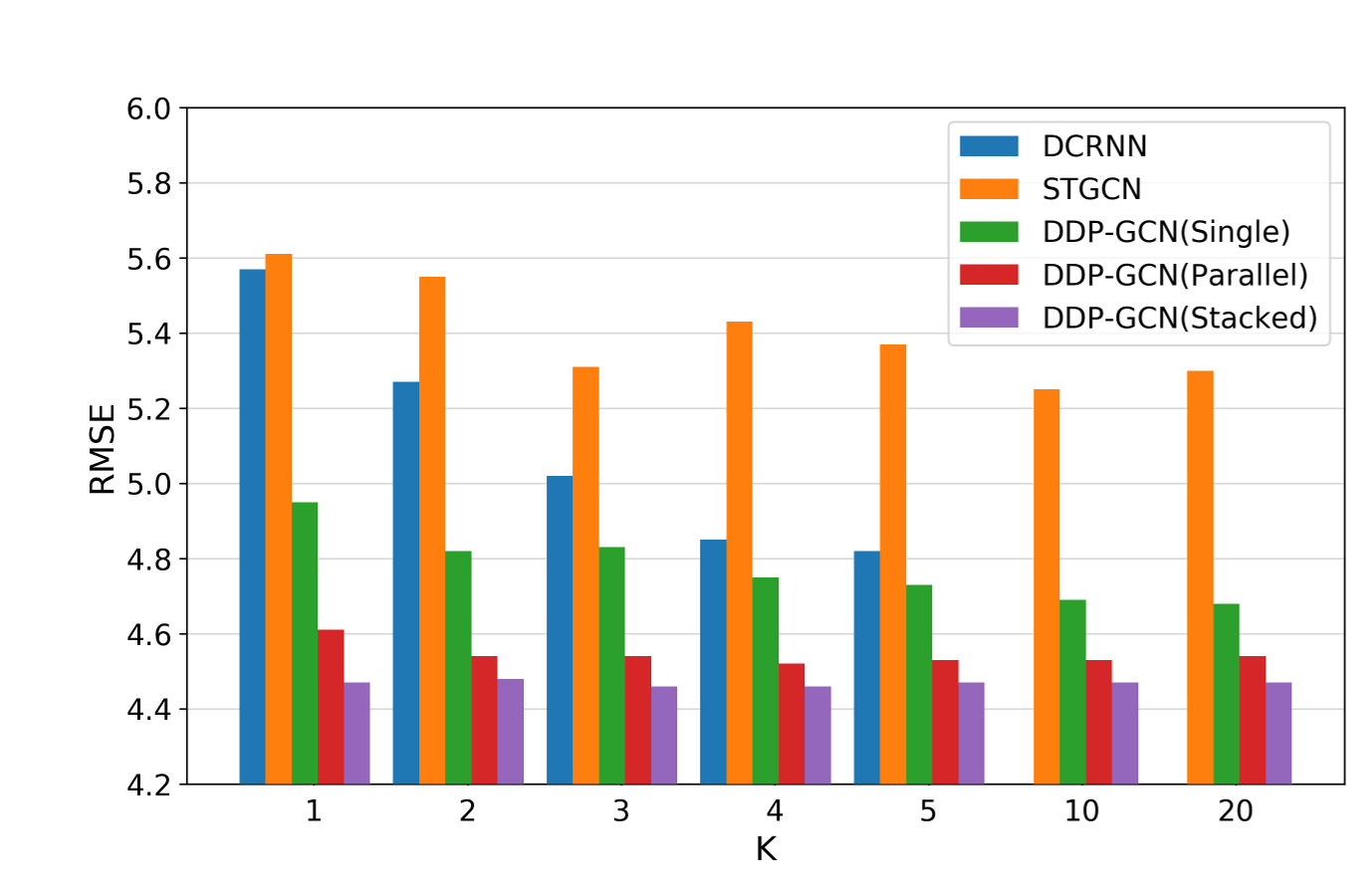}
    }
    \subfloat[Urban2]{
        \includegraphics[width=0.5\textwidth]{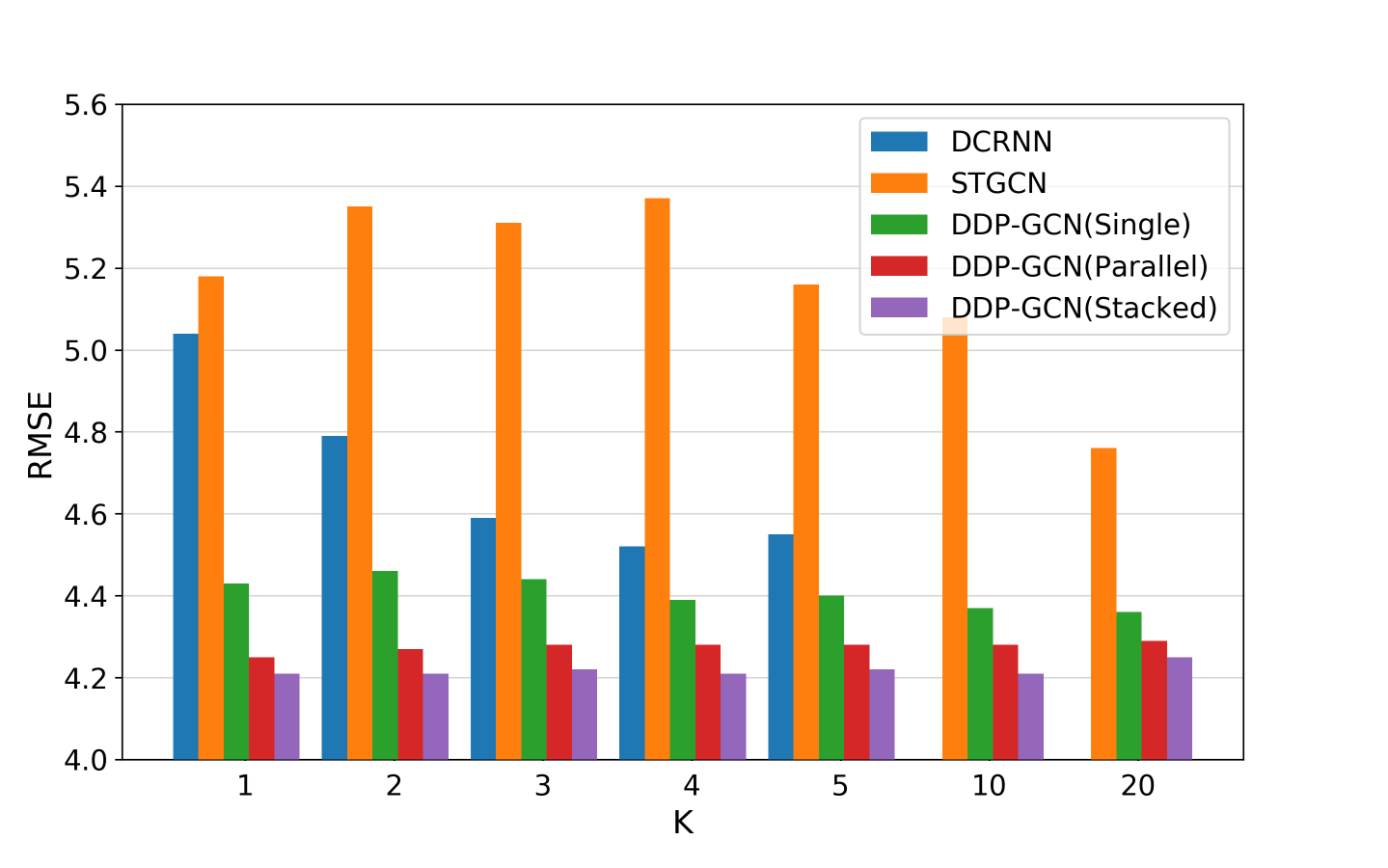}
    }
    \caption{
    Performance with $K$-hops. DDP-GCN(Stacked) can outperform the previously state-of-the-art GCN methods even when they utilize $K$-hops. For DCRNN, the results for $K>5$ were not available due to the long and unstable training process.} 
    \label{fig:large-hops}
\end{figure*}

First, we performed an ablation test where one of the graph elements was chosen and the element was completely removed from DDP-GCN(Stacked). The results are shown in Table \ref{table:ablation}(a).
Removing \textit{Positional relationship}($\left\{ W_D\odot W_P \right\}$) incurred the largest degradation and removing \textit{Distance}($W_D$) the least.
Secondly, we considered replacing each graph element with a distance ($W_D$) element, instead of completely removing the element. The results are shown in Table \ref{table:ablation}(b),
and it can be observed that positional relationship is again the most important factor. 
These results indicate that distance only element is not the most important graph element to reflect complex spatial relationships. On the other hand, direction and positional relationship should be fed into the network for more accurate forecasting.

We have also performed ablation test for partition filters. We have directly applied $W_\theta$ and $W_D$ instead of $T_M(W_\theta)$ or $T_M'(W_D)$. As shown in Table \ref{table:ablation-partiton}, our model performed the best when the partition filters were utilized.

Finally, we have examined the effect of using a larger number of hops. For the investigation, we have used $K$-polynomial ChebNet\cite{defferrard2016} instead of the 1stChebNet. $K$-hops were applied to the distance ($W_D$) element only. As shown in Figure \ref{fig:large-hops}, DDP-GCN(Stacked) outperformed the other GCN models even when they used a large $K$ value. Moreover, the performance of DDP-GCN(Stacked) was not improved for larger $K$. The results indicate that the stacked elements are already performing better than $K$-hops and the stacking of spatial elements outperforms simple $K$-hops. 
\section{Discussion and Conclusion}
\label{conclusion}

In this work, we proposed a new traffic speed forecasting network utilizing three spatial dependencies, namely distance, direction, and positional relationships. 
Our model is constructed using multi-graph convolution with spatial graph elements as the building blocks. We also introduced partition filters that can be used to sub-divide each spatial graph element into multiple components with similar characteristics.
We have evaluated our models together with popular baseline models, and have found improvements especially for long-term forecasting of highly complex urban networks. 

In our study, we have focused on developing graph convolutional networks that can specifically handle spatial dependencies of traffic datasets. Temporal dependencies are also utilized by the networks, but the input was limited to be a consecutive time sequence and thus daily periodicity and weekly periodicity were not utilized. While periodicity is an important temporal characteristics to take advantage of, it requires a careful thinking to modify graph convolutional networks because simply increasing the input sequence size will most likely cause a performance degradation instead of a performance enhancement. This is in contrast to a simple model like Historical Average that can take advantage of weekly periodicity essentially at zero computational overhead. For our highly complex urban networks, weekly periodicity turned out to have a significant forecasting power and thus the performance of Historical Average turned out to be competitive. 
As a future work, it will be promising to consider other ways of constructing inputs in time dimension and to develop graph convolutional networks that can fully exploit the periodicity.

In our study, the three spatial dependencies turned out to be useful for traffic forecasting of a crowded city. The proposed DDP-GCN, however, is a deep learning model and it is unclear exactly how the three spatial dependencies contribute to the performance improvement. This remains as a limitation of our work. To address this issue, a possible future work is to investigate many traffic datasets simultaneously. Such an investigation might be able to pinpoint what aspects of traffic characteristics make direction and positional relationship information helpful for traffic speed forecasting.

\section{Acknowledgment}
This work was supported by a National Research Foundation of Korea (NRF) grant funded by the Korea government (MSIT) (No.~NRF-2020R1A2C2007139), 
Electronics and Telecommunications Research Institute (ETRI) grant funded by the Korean government [21ZR1100, A Study of Hyper-Connected Thinking Internet Technology by autonomous connecting, controlling and evolving ways],
and in part by IITP grant funded by the Korea government (No. 2021-0-01343, Artificial Intelligence Graduate School Program (Seoul National University)). 


\bibliographystyle{elsarticle-num}
\bibliography{ref}

\end{document}